\PassOptionsToPackage{unicode}{hyperref}
\PassOptionsToPackage{hyphens}{url}
\PassOptionsToPackage{dvipsnames,svgnames,x11names}{xcolor}
\documentclass[
]{article}
\usepackage{amsmath,amssymb}
\usepackage{iftex}
\ifPDFTeX
  \usepackage[T1]{fontenc}
  \usepackage[utf8]{inputenc}
  \usepackage{textcomp} 
\else 
  \usepackage{unicode-math} 
  \defaultfontfeatures{Scale=MatchLowercase}
  \defaultfontfeatures[\rmfamily]{Ligatures=TeX,Scale=1}
\fi
\usepackage{lmodern}
\ifPDFTeX\else
\fi
\IfFileExists{upquote.sty}{\usepackage{upquote}}{}
\IfFileExists{microtype.sty}{
  \usepackage[]{microtype}
  \UseMicrotypeSet[protrusion]{basicmath} 
}{}
\makeatletter
\@ifundefined{KOMAClassName}{
  \IfFileExists{parskip.sty}{%
    \usepackage{parskip}
  }{
    \setlength{\parindent}{0pt}
    \setlength{\parskip}{6pt plus 2pt minus 1pt}}
}{
  \KOMAoptions{parskip=half}}
\makeatother
\usepackage{xcolor}
\usepackage{longtable,booktabs,array}
\usepackage{calc} 
\usepackage{etoolbox}
\makeatletter
\patchcmd\longtable{\par}{\if@noskipsec\mbox{}\fi\par}{}{}
\makeatother
\IfFileExists{footnotehyper.sty}{\usepackage{footnotehyper}}{\usepackage{footnote}}
\makesavenoteenv{longtable}
\providecommand{\tightlist}{%
  \setlength{\itemsep}{0pt}\setlength{\parskip}{0pt}}
\usepackage[T1]{fontenc}
\usepackage{float}
\DeclareUnicodeCharacter{03BA}{\ensuremath{\kappa}}
\DeclareUnicodeCharacter{2248}{\ensuremath{\approx}}
\DeclareUnicodeCharacter{2265}{\ensuremath{\geq}}
\DeclareUnicodeCharacter{2264}{\ensuremath{\leq}}
\DeclareUnicodeCharacter{2212}{\ensuremath{-}}
\usepackage{tikz}
\usetikzlibrary{positioning,arrows.meta}
\ifLuaTeX
  \usepackage{selnolig}  
\fi
\IfFileExists{bookmark.sty}{\usepackage{bookmark}}{\usepackage{hyperref}}
\IfFileExists{xurl.sty}{\usepackage{xurl}}{} 
\hypersetup{
  pdftitle={The Test Oracle Problem in Synthetic LLM-as-Judge Corpora: Disappearance, Distortion and a Validation Protocol},
  pdfkeywords={LLM-as-a-judge, test oracle problem, metamorphic testing,
synthetic data, dataset artifacts, multilingual evaluation,
reproducibility},
  colorlinks=true,
  linkcolor={Blue},
  filecolor={Maroon},
  citecolor={Blue},
  urlcolor={Blue},
  pdfcreator={LaTeX via pandoc}}

\title{The Test Oracle Problem in Synthetic LLM-as-Judge Corpora:
Disappearance, Distortion and a Validation Protocol}
\author{Serkan Ballı\\ Department of Software Engineering,\\ Mehmet Akif Ersoy University, Türkiye\\ \texttt{serkanballi@mehmetakif.edu.tr}}
\date{}

\begin{document}
\maketitle
\begin{abstract}
Studies of bias in LLM-as-judge systems typically build synthetic
corpora by prompting an LLM to generate a hallucinated answer to pair
with a factual one, then presenting both to a judge. We report a case in
which this generation step silently failed, and use it to argue that the
failure mode is structural rather than incidental. In a multilingual
(Turkish/English) faithfulness-judgment corpus, a decoding-budget
parameter shared between judging and generation calls truncated one
producer's hallucinated answers to a few words. The resulting items
produced a large, statistically robust effect: a 32-point cross-lingual
collapse in one judge's selection accuracy, replicated from N=50 to
N=500, explained by a three-layer mechanistic account, and confirmed by
a controlled producer-swap experiment, none of which was real. The
effect vanished to ceiling once the shared parameter was corrected, and
only manual reading of the raw generations, not any aggregate
statistical check, exposed the fault. A second measured bias
(Markdown-formatting preference) was not fabricated but distorted by the
same fault, its magnitude and in one case its sign shifting with
stimulus length, a mode aggregate metrics cannot distinguish from the
first. We frame the underlying vulnerability using the test oracle
problem: corpora whose negative examples are LLM-generated carry no
mechanical way to verify item integrity, while corpora built by
deterministic perturbation of a gold answer carry an item-level oracle
for free. A positive control supports this claim directly: an analogous
fault injected into a minimal perturbation-based corpus is caught with
100\% accuracy by a zero-cost, zero-human gold-to-negative string
comparison. We close with a validation protocol, derived from our own
case, for analysts working in the oracle-less regime that we argue
describes most contemporary multilingual LLM-as-judge corpora.
\end{abstract}

\hypertarget{introduction}{%
\section{1. Introduction}\label{introduction}}

Large language models are now widely used as automatic judges of other
models' outputs, and the biases those judges carry (toward verbose
answers, toward Markdown formatting, toward their own family of models)
are themselves an active object of study. The standard experimental
response, building on a broad literature on hallucination detection and
measurement (\hyperlink{ref:alansari}{Alansari \& Luqman, 2026}), is to
construct a synthetic corpus in which each item pairs a \emph{correct}
answer with a \emph{hallucinated} one, present both to a judge, and
measure what the judge does. Almost without exception, the hallucinated
half of each item is itself produced by an LLM, prompted to ``generate a
fluent but incorrect response.'' This generation step is treated as
infrastructure: something that happens once, upstream of the
measurements that matter, and that can be trusted to behave as intended.

In this paper we report a case in which it did not, and we argue that
the failure is not an anecdote about one careless pipeline but a
structural property of the design that produced it. While building a
multilingual judge-bias corpus, we shared a single decoding-budget
parameter between the judging calls and the generation calls. The budget
was adequate for the judges' one-token outputs and catastrophic for the
generators' paragraph-length ones, silently truncating the hallucinated
half of every Turkish item produced by one of two producers. The
truncated items were then judged by three models, and the judgments,
aggregated, described a striking and statistically robust effect: a
32-point collapse in one judge's selection accuracy on Turkish, absent
in English, replicated from N=50 to N=500, supported by a three-layer
mechanistic account, and confirmed by a controlled producer-swap
experiment. None of it was real. Every standard robustness procedure we
applied made the fabricated effect more convincing, not less, and none
of them located the fault. Only reading the raw generated text by hand
did.

We could report this as a cautionary tale and stop. We think that would
understate it. The reason four well-designed robustness checks failed is
not that they were performed carelessly but that none of them is
\emph{designed} to detect a fault in the stimuli themselves: they all
operate on aggregate judge behavior, which is downstream of, and
therefore blind to, item-level degeneration. Whether such a fault
\emph{can} be detected cheaply is a property of how the items were
produced, and that property has a name in software engineering. It is
the \textbf{test oracle problem}: the question of whether a cheap,
mechanical procedure exists for deciding whether a given output is
correct (\hyperlink{ref:barr}{Barr et al., 2015}).

Viewed through this lens, the two ways of building negative examples for
a judge-bias corpus are not symmetric. A \emph{mechanical-perturbation}
design, which derives each hallucinated answer from a gold answer by a
deterministic transform, carries its own item-level oracle for free: the
transform is a relation between gold and negative, checkable by a string
comparison, in the spirit of metamorphic testing
(\hyperlink{ref:chen1998}{Chen et al., 1998}). An
\emph{LLM-generated-negative} design, which samples the hallucinated
answer from a model, carries no such oracle: a generated hallucination
that is degenerate is surface-identical to one that is merely short, and
telling them apart is a semantic act an oracle is meant to spare us. The
fault we document is precisely the kind the second design cannot
mechanize and the first can.

This reframing turns our pipeline failure into a controlled comparison.
We ask: \emph{in an oracle-less generation design, can the standard
aggregate-robustness toolkit surface a stimulus-level fault?} Our
evidence says no, as four separate checks did not, and we show that the
same fault class, injected into an oracle-bearing design, is caught
mechanically and instantly, at no API or human cost. We then ask what an
analyst working in an oracle-less regime \emph{can} do, and our answer
is the one step that worked for us: read the raw items, by hand, before
computing statistics.

We make three contributions.

\begin{enumerate}
\def\labelenumi{\arabic{enumi}.}
\item
  \textbf{An existence proof.} We document a concrete, A/B-verified case
  in which a generation fault in an LLM-generated-negative corpus
  manufactured a cross-lingual bias effect that survived replication, a
  large sample, a converging mechanistic account, and a controlled
  manipulation, and was exposed only by manual reading of the raw
  stimuli (\hyperlink{sec:4.1}{Section~4.1}). We further show, on the
  corrected data, that the fault corrupted the measured biases in two
  categorically different ways, fabricating one effect outright while
  bending a real one (\hyperlink{sec:4.2}{Section~4.2}), and that
  aggregate metrics cannot distinguish the two.
\item
  \textbf{A positive control.} We inject the closest mechanical analogue
  of our fault into a minimal oracle-bearing (mechanical-perturbation)
  corpus built from the same source stream, and show it is caught at
  100\% accuracy by a gold-to-negative string comparison with zero API
  calls and zero human reading (\hyperlink{sec:4.3}{Section~4.3}). This
  converts the claim ``one design is auditable and the other is not''
  from an argument into a demonstration, and locates the asymmetry in
  test-oracle theory rather than in a bespoke taxonomy.
\item
  \textbf{A validation protocol.} For analysts who must work in the
  oracle-less regime, which includes most contemporary multilingual
  judge work, we prescribe a concrete, cheap check to run before any
  aggregate statistic is computed: manually read fifteen to twenty raw
  generated items per generation condition, and report their length
  distribution and degeneration rate (\hyperlink{sec:5}{Section~5}). We
  derive the specific thresholds from our own case rather than asserting
  them as universal, and we discuss the conditions under which they
  should generalize.
\end{enumerate}

We are deliberate about what this paper is not. It is not the discovery
of a new judge bias: the one bias our corrupted pipeline made dramatic
was an artifact, and the one it left intact is a known formatting
preference. It is not a census of which existing corpora are affected:
we have audited only our own pipeline, and we cite candidate members of
the affected class without verifying them. And it is not a claim that
mechanical-perturbation designs are otherwise superior, only that their
faults are findable. What it is, we hope, is a short, honest, and
reproducible argument that in synthetic judge-bias research the
integrity of the stimuli bounds the integrity of every measurement built
on them, that the relevant design choices make this bound more or less
mechanically checkable, and that the field's current default design
leaves that check to the analyst's own discipline.

\hypertarget{background-and-related-work}{%
\section{2. Background and Related
Work}\label{background-and-related-work}}

Our argument addresses two readerships, and we treat them in turn.

\textbf{LLM-as-judge bias and synthetic evaluation corpora.} Our work
sits within a recent and fast-growing literature that measures the
biases LLM judges bring to evaluation and that builds the synthetic
corpora such measurements require. A recent comprehensive survey
(\hyperlink{ref:gu2025}{Gu et al., 2025}) catalogs the biases this
literature measures, grouping them into task-agnostic and task-specific
classes; we take that taxonomy as background rather than as our object.
Closest to ours as a structural relative is \emph{Judging the Judges}
(\hyperlink{ref:soumik}{Soumik, 2026}, TMLR under review), a study of
five judges, four bias types, and nine mitigation strategies that
reports style as the dominant bias (0.76--0.92) and position bias as
minimal (at most 0.04). We share neither its judge set, which is fully
disjoint from ours (Gemini 2.5 Pro and Flash, Claude Sonnet 4, GPT-4o,
Llama 3.3-70B versus our DeepSeek, Gemini-flash, and GLM, with only the
Gemini family in common), nor its object: where that work documents
within-language scoring shifts and an inventory of mitigations, we
report an accuracy collapse on a low-resource language that a producer
swap isolates, and our contribution is upstream of bias measurement, on
the integrity of the stimuli the measurement is run on. Soumik's
\emph{Judging the Judges} shares its title with an unrelated study of
position bias (\hyperlink{ref:shi}{Shi et al., 2025}), which we cite
separately in \hyperlink{sec:3.4}{Section~3.4}; the coincidence of title
is just that. BabelJudge (\hyperlink{ref:shreyas}{Shreyas KC, 2026})
documents a cross-lingual reliability gap across languages and agent
trajectories, naming a phenomenon adjacent to the one our corrupted
pipeline manufactured; we defer its perturbation-based construction to
\hyperlink{sec:4.3}{Section~4.3}, where it is central. A second strand
addresses bias through aggregation: PoLL
(\hyperlink{ref:verga}{Verga et al., 2024}) shows that a diverse panel
of small judges can outperform a single large one, and UDA
(\hyperlink{ref:zhang}{Zhang et al., 2026}) reports heterogeneous,
self-favoring judge biases that partial aggregation can absorb; against
these, \hyperlink{ref:ma}{Ma et al. (2025)} caution that multi-agent
judging can amplify bias rather than reduce it, a counterpoint we return
to in \hyperlink{sec:4.2}{Section~4.2}. Adjacent to our investigation is
Preference Leakage (\hyperlink{ref:li2026}{Li et al., 2026}), which
traces evaluation distortion to the producer-judge relationship. We draw
the comparison to document a false lead honestly: in
\hyperlink{sec:4.1}{Section~4.1} we initially treated a producer-judge
interaction as our central result and distinguished it from the
family-relatedness mechanism that work identifies, on the grounds that
our producer and judge (DeepSeek and GLM) come from different families
and companies; that interaction was later shown to be an artifact of the
same generation fault, and we record the distinction here only to make
clear why an unrelated-family explanation seemed plausible before the
manual read exposed the truncation.

CalibJudge (\hyperlink{ref:wen}{Wen et al., 2026}) we cite only as a
candidate member of the LLM-generated-negative design class: it is an
unrefereed preprint, and we could not confirm whether its negatives are
LLM-generated or drawn from an existing error corpus, so we treat the
broader question of which existing corpora fall on which side of our
oracle line as open rather than settled. Across this literature the
synthetic corpus is treated as infrastructure and the integrity of its
items as assumed; our contribution is upstream of all of it, on the
conditions under which that integrity can be checked at all. As a matter
of genre, the paper belongs to the dataset- and annotation-artifact
tradition in NLP, in which the data and the pipeline that produced it,
rather than a novel finding extracted from it, are themselves the
contribution.

\textbf{The test oracle problem and metamorphic testing.} The structural
argument we make draws its vocabulary from software testing. The test
oracle problem (\hyperlink{ref:barr}{Barr et al., 2015}) is the
observation that for many systems no cheap, mechanical procedure exists
for deciding whether an output is correct, and that the only available
check is to redo the work the system did, by hand or by expensive
semantic judgment. The field's response, where a full oracle is
unavailable, is metamorphic testing
(\hyperlink{ref:chen1998}{Chen et al., 1998}; its development and
applications are surveyed in
\hyperlink{ref:chen2018}{Chen et al., 2018}): rather than asserting that
an output is correct in absolute terms, one defines a relation the
output should satisfy under a transformation of the input, and checks
the relation, which is typically cheap and mechanical even when absolute
correctness is not. Our use of these ideas is narrow and specific. We do
not apply metamorphic testing to the judges; we apply its logic to the
generation design, observing that a corpus whose negatives are derived
from a gold answer by a deterministic transform carries a metamorphic
relation (gold-to-negative) for free, and that checking it is a
constant-time string comparison, whereas a corpus whose negatives are
sampled from a model carries no such relation. To our knowledge the
oracle problem has not been named as the structural distinction between
these two corpus-construction designs in the LLM-as-judge literature,
and naming it is the contribution of \hyperlink{sec:4.3}{Section~4.3}.
The two bodies of work address different objects, the bias of the judge
and the integrity of the stimulus, and we keep them separate because
conflating them would obscure both.

\hypertarget{method}{%
\section{3. Method}\label{method}}

We describe the materials and procedures common to the findings. All
measurements are reproducible from a single cached log of roughly 17,800
LLM calls and the accompanying scripts; no result in the paper requires
additional API calls. The cache and scripts are released as a public
deposit (\hyperlink{sec:3.7}{Section~3.7}).

\hypertarget{judges}{%
\subsection{3.1 Judges}\label{judges}}

Three judges, drawn from three model families to avoid within-family
confounds: \textbf{DeepSeek} (DeepSeek-V4-Flash via the official API),
\textbf{Gemini} (gemini-3.5-flash, with the thinking trace disabled via
\texttt{thinking\_budget=0}, so that internal reasoning does not consume
the output token budget), and \textbf{GLM} (glm-5.2, accessed via z.ai's
Anthropic-compatible API). The three judges serve both as evaluators of
the items and, for two of them, as producers of the items, which is the
producer-judge interaction we examine in
\hyperlink{sec:4.1}{Section~4.1}.

\hypertarget{data}{%
\subsection{3.2 Data}\label{data}}

Two item sources, one per language. \textbf{Turkish} items are
generated, not retrieved: for each of N contexts drawn from the Turkish
split of wiki40b (\hyperlink{ref:wiki40b}{wiki40b, 2020}; first 80
words, length \textgreater{} 60 chars), one producer LLM is prompted to
emit a \emph{factual} answer (``a short answer faithful to the
context'') and a \emph{hallucinated} answer (``a contradictory/wrong
answer''), producing a
\texttt{\{context,\ question,\ factual,\ halluc\}} quadruple. The
default question is the open ``What do you think about this text?'',
matching a faithfulness- rather than factoid-reading task. Two producers
are used in the controlled comparison of
\hyperlink{sec:4.1}{Section~4.1}: DeepSeek and Gemini. \textbf{English}
items are taken verbatim from HaluEval
(\hyperlink{ref:li2023}{Li et al., 2023}, qa split), whose
\texttt{right\_answer} and \texttt{hallucinated\_answer} fields provide
ready-made pairs; HaluEval is a public benchmark and its items were
\emph{not} re-generated by us.

\hypertarget{the-ab-design-corrupted-versus-corrected-pipeline}{%
\subsection{3.3 The A/B design: corrupted versus corrected
pipeline}\label{the-ab-design-corrupted-versus-corrected-pipeline}}

The paper's central comparison is between two versions of the same
generation pipeline, applied to the same source contexts.

\begin{itemize}
\tightlist
\item
  \textbf{Corrupted pipeline} (the configuration under which the
  artifact was originally measured). Generation and judging share a
  single utility function and, with it, a single \texttt{max\_tokens}
  budget of 5, sufficient for a judge's ``A''/``B'' output but
  insufficient for a paragraph-length generation, which is consequently
  truncated. The Gemini generation path, implemented separately, carries
  no explicit cap and is unaffected. Generation prompts do not forbid
  conversational openers.
\item
  \textbf{Corrected pipeline.} Generation and judging budgets are
  separated (\texttt{max\_tokens=300} for generation, 5 for judging).
  The Gemini thinking trace is disabled. The generation prompt is
  tightened to forbid conversational wrappers (``write only the
  information; do not use introductory phrases such as `of course'\,'').
  Cached generation calls are namespaced (\texttt{gen300::}) so they
  cannot collide with the corrupted-pipeline cache.
\end{itemize}

The two pipelines are applied to the \emph{same} source contexts, so any
difference in measured bias is attributable to the pipeline, not to the
data. This is the A/B comparison that grounds
\hyperlink{sec:4.1}{Section~4.1}.

\hypertarget{sec:3.4}{%
\subsection{3.4 Measured biases}\label{sec:3.4}}

For each item and judge we measure four quantities.

\begin{itemize}
\tightlist
\item
  \textbf{Style.} The judge is presented with the factual answer in two
  formats, plain and prefixed with a Markdown bullet (\texttt{-\ X}), in
  fixed A-first order, and asked which is more faithful. The style score
  is the fraction of items on which the Markdown version is chosen. (We
  note, and examine in \hyperlink{sec:4.2}{Section~4.2}, that this
  quantity depends on the length of the string to which the bullet is
  attached.)
\item
  \textbf{Position.} The judge sees factual-versus-hallucinated in both
  A-first and B-first order; the difference in factual-selection rate is
  the position bias (an independently studied quantity;
  cf.~\hyperlink{ref:shi}{Shi et al., 2025}), where a value near 0
  indicates the judge is order-insensitive.
\item
  \textbf{Selection.} The judge sees the same
  factual-versus-hallucinated pair labeled A/B in one call and 1/2 in
  another; the rate at which the two calls agree is the
  selection-consistency score, where low values flag label-driven rather
  than content-driven choice.
\item
  \textbf{Factual accuracy.} On the factual-versus-hallucinated task,
  the rate at which the judge selects the factual answer; this is the
  primary faithfulness measure.
\end{itemize}

Inter-judge agreement is reported as Fleiss' κ across the three judges.
We additionally report a three-judge majority-vote accuracy as an
incidental check on aggregation.

\hypertarget{item-level-audit-and-length-stratification}{%
\subsection{3.5 Item-level audit and length
stratification}\label{item-level-audit-and-length-stratification}}

Two offline analyses, both run from the cache with no API calls. The
\textbf{contamination audit} (\hyperlink{sec:4.1}{Section~4.1})
classifies each generated hallucination as a verbatim copy of the
factual field, a truncated conversational filler (opener in a fixed
gazetteer), a short non-filler fragment (at most 3 words), or a valid
hallucination, and reports the rate of each plus the mean word length of
factual and hallucinated fields. The \textbf{style-by-length
stratification} (\hyperlink{sec:4.2}{Section~4.2}) recomputes each
judge's style score after binning items by factual-field word length,
across the corrupted Turkish set, the unaffected Gemini-produced Turkish
set, and the English HaluEval set, to test whether the style score is
invariant to the length of the string it is measured on.

\hypertarget{positive-control-procedure-section-4.3}{%
\subsection{3.6 Positive-control procedure (Section
4.3)}\label{positive-control-procedure-section-4.3}}

To test whether the fault class of \hyperlink{sec:4.1}{Section~4.1} is
mechanically catchable in an oracle-bearing design, we build a minimal
mechanical-perturbation corpus from the same wiki40b Turkish stream:
fifty sentences containing numerals, each perturbed by replacing its
first numeral. We inject a deliberate analogue of our fault via per-item
independent Bernoulli draws at a 30\% rate (frozen by seed 42; the
realized count was 20 no-ops, 40\% of the 50 items). On a no-op draw the
substitution is silently skipped, emitting a negative identical to its
gold. The oracle is a gold-to-negative string comparison: a perturbation
that leaves the string unchanged is flagged instantly. We report the
oracle's accuracy at separating genuine perturbations from no-ops, and
its cost in API calls and human reading. This procedure uses no LLM and
no human annotator; it is included precisely to show that the relevant
check is cheap \emph{when the design admits it}.

\hypertarget{sec:3.7}{%
\subsection{3.7 Reproducibility}\label{sec:3.7}}

Every LLM call in the study is logged in a single cache keyed by judge
and prompt, so that every reported number can be regenerated by
replaying the analyses against the cache with no new API expenditure.
One environmental note is worth recording because it is itself a member
of the fault class this paper describes: Gemini flash-latest is a
thinking model, and failing to disable its thinking trace silently
consumes the output token budget, the same kind of silent truncation we
study. A HuggingFace streaming-shutdown warning is emitted during
dataset teardown; it fires after all results are logged and does not
affect any reported number. Both producer aliases were stable throughout
data collection (July 2026): deepseek-chat resolved to DeepSeek-V4-Flash
(released April 2026) and gemini-flash-latest to gemini-3.5-flash
(generally available May 2026); neither alias changed during the study,
so the producer-judge results are not confounded by silent model-version
drift, the very fault class this paper warns against.

\textbf{Code availability.} The cached LLM-call log and all analysis
scripts are deposited at OSF
(\href{https://doi.org/10.17605/OSF.IO/RS2B8}{10.17605/OSF.IO/RS2B8}).
The deposit comprises five self-contained scripts, one per analysis: the
corrupted-pipeline bias measurement, the corrected-pipeline measurement,
the item-level contamination audit, the style-by-length stratification,
and the mechanical-perturbation positive control. The positive control
and the two offline audits require no API calls; only the two
bias-measurement scripts replay vendor APIs, and they too can be run
cache-only. No external services beyond the three vendor APIs are used.

\hypertarget{findings}{%
\section{4. Findings}\label{findings}}

We present the findings in the order that makes the structural argument
legible: the anatomy of the artifact and why standard checks missed it
(4.1), the two qualitatively different ways the fault distorted the
measured biases (4.2), and the oracle asymmetry that explains the
difference in detectability (4.3).

\hypertarget{sec:4.1}{%
\subsection{4.1 Anatomy of an Artifact}\label{sec:4.1}}

We present this account in the order the evidence accumulated, rather
than moving directly to the corrected result, because the sequence is
itself the central claim of this section: every standard robustness
procedure we applied made the finding more convincing, not less, and
none of them located the fault. Only reading raw generated text by hand
did.

\textbf{An apparently robust signal.} In an initial pass over N=500
Turkish faithfulness-judgment items, GLM's selection accuracy on
hallucination pairs produced by DeepSeek stood at approximately 0.64,
against 0.96 on the equivalent English benchmark (HaluEval), a 32-point
gap we initially read as cross-lingual degradation. Gemini and DeepSeek,
judging the same items, scored 0.99 and 0.85 respectively, an ordering
that matched each model's independently measured baseline competence in
Turkish. A 95\% confidence interval placed the estimate at {[}0.60,
0.68{]}, several standard errors from chance; Fleiss' κ across the three
judges was 0.041.

\textbf{Replication.} The same pattern had first appeared at N=50
(selection accuracy 0.54, with a correspondingly wide confidence
interval), and scaling to N=500 tightened the estimate without changing
its direction. We treated this as evidence against measurement noise,
correctly, in the narrow sense that the underlying number was indeed
stable, though a different, undetected problem remained.

\textbf{A three-layer mechanism account.} We then built a causal
explanation in three stages. First, we compared tokenization
fragmentation across the three judges on the same Turkish text and found
GLM to have the lowest fragmentation of the three, an inverse
relationship we took as evidence against a tokenizer-based account;
tokenization-fairness analyses
(\hyperlink{ref:petrov}{Petrov et al., 2023}) predict that higher
fragmentation should disadvantage a model, so the fact that the
least-fragmented judge was the most affected argues against, not for,
such an account. Second, we turned to the judges' technical reports: the
GLM-4 technical report (\hyperlink{ref:glm4}{Team GLM, 2024}) documents
a roughly ten-trillion-token pretraining corpus dominated by Chinese and
English, with a comparatively small supplementary corpus spanning 24
additional languages, while the GLM-5 report
(\hyperlink{ref:glm5}{GLM-5 Team, 2026}) describes a benchmark suite
oriented entirely around agentic and coding tasks, with no
natural-language multilingual evaluation reported; we read this
family-level shift as indirect support for a
thinning-multilingual-exposure account. Third, following a controlled
producer-swap experiment (below) that appeared to implicate the specific
generator rather than the language itself, a post-hoc comparison of
lexical overlap between hallucinated spans and source documents found no
difference between GLM's correct and incorrect calls (0.19 vs.~0.18),
evidence against a ``subtle hallucination'' account, while a length
comparison revealed DeepSeek's Turkish generations averaged
approximately two words against roughly forty-five for Gemini. We
reinterpreted the effect through an item-response-theory lens: an
extremely short, low-information stimulus should disproportionately
penalize whichever judge has the thinnest competence margin, which,
given the same competence ordering established above, predicts exactly
the graded severity we observed (Gemini least affected, DeepSeek
intermediate, GLM most affected).

\textbf{A controlled experiment.} To test whether the effect tracked the
producer rather than the language, we constructed a second Turkish item
set using Gemini, rather than DeepSeek, as the hallucination generator,
holding judges and task fixed. GLM's selection accuracy on the
DeepSeek-produced set was 0.61 (N=200 controlled set); on the
Gemini-produced set it was 1.00. We interpreted this as a genuine
producer-judge interaction, distinguishing it explicitly from
family-relatedness-driven preference effects reported elsewhere
(\hyperlink{ref:li2026}{Li et al., 2026}), and treated it as the paper's
central, sharpened contribution.

\textbf{None of it was correct.} Every step above increased our
confidence in a real, mechanistically explicable effect. None of them
was.

\textbf{What twenty transcripts showed.} A manual reading of the
DeepSeek-generated Turkish items, the one step none of the preceding
analyses had included, found that 81.0\% began with a truncated
conversational filler (``Tabii ki, iş\ldots{}'', ``Elbette,
ver\ldots{}''), 12.2\% consisted of the hallucinated field reproducing
the factual field verbatim, and a further 6.8\% were otherwise
degenerate fragments. No item in the sample constituted a well-formed
hallucination. In at least the 12.2\% of cases where the two fields were
identical, the classification task given to the judges was not merely
difficult; it was unanswerable by construction.

\textbf{Locating the fault.} The generation and judging calls in our
pipeline shared a single utility function and, with it, a single
max\_tokens parameter, set to 5, sufficient for a judge's ``A''/``B''
output but not for generating a hallucinated passage, which was
consequently truncated mid-sentence. The Gemini generation calls, made
through a separate code path with no explicit token cap (defaulting to
roughly 4,000 tokens), were unaffected. The asymmetry this created, one
producer silently crippled and the other not, is what had appeared, in
turn, as a language effect and then as a producer effect.

\textbf{Recurrence, and disappearance.} After separating generation and
judging token budgets and re-running the DeepSeek pipeline, generation
validity rose from 0\% to 98.8\% (mean length 20.0 words, against
Gemini's unchanged 46.9), and GLM's selection accuracy on the same
DeepSeek-produced task rose from 0.61 to 1.00. DeepSeek's own apparent
within-producer deficit (factual accuracy 0.88 on its own generations)
resolved to 1.00 under the same correction. No cross-lingual,
producer-specific, or self-referential effect survived the fix.

The lesson we draw is not that any single robustness check was performed
carelessly, but that none of the checks in the current standard toolkit
(larger samples, replication, a converging mechanistic account, a
controlled manipulation) are designed to detect a fault in the stimuli
themselves. Table 1 consolidates the recovery.
\hyperlink{sec:5}{Section~5} formalizes the one check that did.

\begin{table}[H]
\centering
\caption{The corrupted versus corrected generation pipeline on the Turkish DeepSeek-produced set. Every affected quantity recovers under the fix; Gemini-produced items (mean hallucinated length 46.9 words) are unchanged throughout. n/a marks a contamination category not separately re-measured after the fix, as generation validity and mean length already summarize the recovery.}
\begin{tabular}{@{}lll@{}}
\toprule
Quantity & Corrupted & Corrected \\
\midrule
Generation validity & 0.0\% & 98.8\% \\
Mean hallucinated length (words) & 2.1 & 20.0 \\
Contamination, conversational filler & 81.0\% & n/a \\
Contamination, verbatim repeat & 12.2\% & n/a \\
Contamination, short fragment & 6.8\% & n/a \\
GLM selection accuracy (N) & 0.61 (N=200) & 1.00 (N=80) \\
GLM factual accuracy & 0.845 & 0.99 \\
DeepSeek within-producer factual & 0.88 & 1.00 \\
Three-judge majority factual & 0.962 & 1.00 \\
\bottomrule
\end{tabular}
\end{table}

\hypertarget{sec:4.2}{%
\subsection{4.2 Two Modes of Corruption: Disappearance and
Distortion}\label{sec:4.2}}

A reviewer could accept the artifact of \hyperlink{sec:4.1}{Section~4.1}
and still ask a pointed follow-up: \emph{was the selection collapse the
only thing the corrupted pipeline manufactured, or were other measured
biases also distorted?} The corrected data answers this in a way that is
less tidy, and more useful, than a simple all-or-nothing verdict. Two of
the biases we measured behave in qualitatively different ways under the
fix, and the difference is itself the finding.

\textbf{Mode 1, disappearance: the selection collapse was fabricated
wholesale.} After the generation budget was separated from the judging
budget and the conversational wrapper was forbidden, GLM's selection
accuracy on DeepSeek-produced Turkish items rose from 0.61 (N=200,
corrupted) to \textbf{1.00} (N=80, corrected); its factual accuracy rose
from 0.845 to 0.99. DeepSeek's apparent within-producer deficit, factual
accuracy 0.88 on its own generations, which we had briefly entertained
as a self-reference effect, resolved to 1.00 under the same correction.
No cross-lingual, producer-specific, or self-referential effect
survived. This is the clean case: the corrupted pipeline called an
effect into being that had no existence in the corrected data. We do not
treat the null result as a weak replication; we treat it as the defining
confirmation that the effect was a property of the stimuli.

\textbf{Mode 2, distortion: the style preference was real but bent by
the same fault.} Style bias, whether a judge prefers a factual answer
rendered with a Markdown bullet (\texttt{-\ X}) to the same answer in
plain text (\texttt{X}), did \emph{not} disappear, and did \emph{not}
survive unchanged. It varied systematically with the length of the
factual text it was measured on, and it did so in judge-specific ways
that the corrupted pipeline had collapsed into a single misleading
number per language.

To show this we stratified each judge's style score by the word length
of the factual field it was judging, across every condition we had
cached: the corrupted DeepSeek-produced Turkish set (mean 2.1 words),
the unaffected Gemini-produced Turkish set (mean 43 words), and the
English HaluEval benchmark (mean 2.2 words). The three judges fall into
three regimes (Table 2).

\begin{table}[H]
\centering
\caption{Each judge's style score (fraction choosing the Markdown-bulleted version) stratified by factual-field word length. Preferences are genuine signatures, but their magnitude, and for DeepSeek their sign, track length.}
\begin{tabular}{@{}
  >{\raggedright\arraybackslash}p{(\columnwidth - 6\tabcolsep) * \real{0.2500}}
  >{\raggedright\arraybackslash}p{(\columnwidth - 6\tabcolsep) * \real{0.2500}}
  >{\raggedright\arraybackslash}p{(\columnwidth - 6\tabcolsep) * \real{0.2500}}
  >{\raggedright\arraybackslash}p{(\columnwidth - 6\tabcolsep) * \real{0.2500}}@{}}
\toprule
\begin{minipage}[b]{\linewidth}\raggedright
Judge
\end{minipage} & \begin{minipage}[b]{\linewidth}\raggedright
1-to-2-word factuals
\end{minipage} & \begin{minipage}[b]{\linewidth}\raggedright
16+ word factuals
\end{minipage} & \begin{minipage}[b]{\linewidth}\raggedright
Behavior under lengthening
\end{minipage} \\
\midrule
DeepSeek & 0.24 (EN) / 0.43 (TR) & \textbf{0.81} (TR) & Rises, crossing
0.5: plain-preferring when short, Markdown-preferring when long \\
Gemini & 0.00 (EN) / 0.02 (TR) & 0.00 (TR) & Flat near 0;
plain-preferring, length-invariant \\
GLM & 0.90 (EN) / 0.89 (TR) & \textbf{0.65} (TR) & Falls but stays above
0.5; Markdown-preferring, weakened when long \\
\bottomrule
\end{tabular}
\end{table}

(Two-proportion tests: DeepSeek short-versus-long z≈8.7, GLM z≈7.0, both
far beyond noise; Gemini flat by inspection.)

Three things follow from this table. First, the style preferences are
not invented: each judge carries a signature that no length condition
erases (Gemini plain throughout, GLM Markdown throughout). Second, their
\emph{magnitude}, and for DeepSeek their \emph{direction} across the 0.5
line, is a function of how much text the style marker was attached to,
which is a property of the generation pipeline, not of the judge. Third,
and most consequential for our earlier reading: the reason Turkish and
English style scores had looked so similar on the corrupted data was not
that style is language-independent. It was that both the corrupted
Turkish set and the English benchmark happened to contain short factuals
(2.1 and 2.2 words respectively), placing both in the same length
regime. The apparent cross-lingual invariance was a length invariance we
had misread as a language invariance.

\textbf{The English benchmark sits inside the same length regime as the
corrupted data.} This is worth stating plainly, because it is the kind
of fact a careful reader will surface anyway. HaluEval's
\texttt{right\_answer} fields average 2.2 words, essentially the same as
our corrupted DeepSeek Turkish factuals (2.1 words). We used HaluEval as
the English reference \emph{because it is a reputable, established
benchmark}, and we did not re-audit its individual items against the
fifteen-to-twenty-item manual-reading standard we now prescribe for our
own data. The overall style scores we report on it (DeepSeek 0.27,
Gemini 0.00, GLM 0.88) are therefore short-regime scores, directly
comparable to the corrupted Turkish scores and not to the corrected
ones; the table's 1-to-2-word bins are length-stratified subsets and
differ slightly from these whole-set values for the same reason. We flag
this not to retract the English numbers but to name the asymmetry in our
own evidence: the corrected-data claim rests on Turkish alone, and the
English reference was trusted on reputational rather than item-level
grounds. A reviewer who applies our protocol to HaluEval directly would
be doing exactly what we recommend and have not yet done.

\textbf{Framing: two modes, one lesson.} The two biases we measured were
both touched by the generation fault, but in categorically different
ways. The selection collapse was \emph{disappearance}: a fabricated
effect with no corrected counterpart, zero residual. The style
preference was \emph{distortion}: a genuine judge signature whose
measured magnitude and, in one case, sign were functions of a stimulus
property the pipeline controlled. The lesson we draw is narrower and
stronger than the one the first draft of this section offered. It is not
that some synthetic-data biases are artifacts and others are real. It is
that, under a generation design whose item quality is hostage to an
unaudited inference call, \emph{no} measured bias can be safely
partitioned into ``real'' and ``spurious'' from aggregate metrics alone,
because the partition itself depends on stimulus properties the metrics
do not expose. \hyperlink{sec:5}{Section~5} prescribes the one check
that does.

\textbf{An incidental note on aggregation.} On the corrupted Turkish
data the three-judge majority recovered a factual accuracy of 0.962
against GLM's individual 0.642. We flag this without theory: it is
consistent with the standard defense of jury aggregation
(\hyperlink{ref:verga}{Verga et al., 2024};
\hyperlink{ref:zhang}{Zhang et al., 2026}), and also with the less
comforting possibility, foreshadowed by
\hyperlink{ref:ma}{Ma et al. (2025)} finding that multi-agent judging
can amplify bias rather than reduce it, that a majority can hide a
stimulus-level fault rather than reveal it. The corrected-data majority
(1.00) settles the practical question but not the methodological one.

\hypertarget{sec:4.3}{%
\subsection{4.3 The Oracle Asymmetry}\label{sec:4.3}}

The bug of \hyperlink{sec:4.1}{Section~4.1} is ordinary. A single shared
parameter, set for one purpose and inherited by another, is a mistake
that could appear in any codebase, in any field; there is nothing
structural about it, and if this section merely said ``we made a coding
error and here is how to avoid it,'' it would not belong in a paper.
What \emph{is} structural is not the bug but its \textbf{detectability},
and detectability, in this case, is governed by a concept software
engineering has studied for decades: the \emph{test oracle problem}.

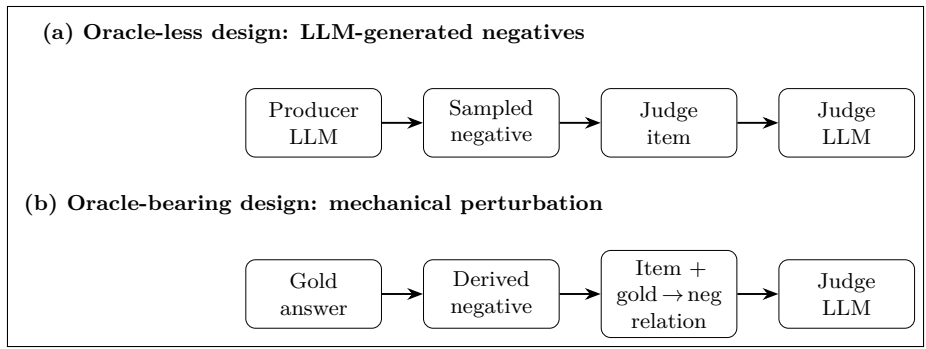
\begin{figure}[t]
\centering
\fbox{\resizebox{\dimexpr\textwidth-2\fboxsep-2\fboxrule\relax}{!}{%
\begin{tikzpicture}[
  box/.style={draw, rounded corners, align=center, inner sep=4pt, font=\small, minimum height=10mm, minimum width=18mm, text width=17mm},
  hdr/.style={align=left, font=\small\bfseries},
  arr/.style={-Stealth, thick},
  note/.style={align=center, font=\footnotesize\itshape, text width=70mm}
]
\node[hdr] at (0,4.7) {(a) Oracle-less design: LLM-generated negatives};
\node[box] (prod) at (0,3.4) {Producer\\LLM};
\node[box, right=6mm of prod] (hall) {Sampled\\negative};
\node[box, right=6mm of hall] (it) {Judge\\item};
\node[box, right=6mm of it] (jd) {Judge\\LLM};
\draw[arr] (prod) -- (hall); \draw[arr] (hall) -- (it); \draw[arr] (it) -- (jd);
\node[hdr] at (0,2.2) {(b) Oracle-bearing design: mechanical perturbation};
\node[box] (gold) at (0,0.9) {Gold\\answer};
\node[box, right=6mm of gold] (der) {Derived\\negative};
\node[box, right=6mm of der] (itb) {Item +\\gold\,$\to$\,neg\\relation};
\node[box, right=6mm of itb] (jdb) {Judge\\LLM};
\draw[arr] (gold) -- (der); \draw[arr] (der) -- (itb); \draw[arr] (itb) -- (jdb);
\end{tikzpicture}%
}}
\caption{The two generation designs sit on opposite sides of the oracle line. In (a) the hallucinated answer is sampled from a producer LLM, so no mechanical check distinguishes a degenerate item from a merely short one. In (b) the hallucinated answer is derived from a gold answer by a deterministic perturbation, so the gold-to-negative relation is a free item-level oracle.}
\label{fig:designs}
\end{figure}

\textbf{The oracle problem, briefly.} A test oracle is any mechanism
that decides, mechanically, whether a program output is correct. The
oracle problem is the observation that for many systems no cheap oracle
exists: the only way to know whether an output is right is to do the
same work the system did, by hand or by expensive semantic judgment
(\hyperlink{ref:barr}{Barr et al., 2015}). Where a cheap oracle exists,
item-level faults are caught trivially and at scale; where one does not,
they are not. We argue that the two designs for producing negative
examples in synthetic judge-bias corpora sit on opposite sides of this
line, and that this, not the relative frequency of bugs, is the
asymmetry that matters. Figure \ref{fig:designs} depicts the two designs
side by side.

\textbf{Mechanical perturbation carries an item-level oracle.} In a
corpus built by \emph{gold-labeling by degradation} (take a correct
answer, apply a deterministic transformation such as an entity swap, a
number change, or a negation, and emit the result as the hallucinated
counterpart; cf.~\hyperlink{ref:shreyas}{BabelJudge, 2026}), every item
comes with its own oracle for free. The transformation is a relation
between gold and negative, and verifying that the relation holds is a
string comparison: take the gold, apply the transform, and check that
the result equals the emitted negative. This is metamorphic testing in
its canonical form (\hyperlink{ref:chen1998}{Chen et al., 1998}): one
does not claim the gold is \emph{absolutely} correct, only that the
gold-to-negative \emph{relation} holds, and that relation is checkable
in O(1) with no semantic judgment. A fault in the perturbation script,
such as a regex that misses or a gazetteer entry that silently returns
the same entity, is exposed the instant it produces a negative that does
not differ from its gold, because the oracle is sitting right there in
the data structure.

\textbf{LLM-generated negatives carry no item-level oracle.} In a corpus
built by prompting an LLM to ``produce a fluent but incorrect response''
(our pipeline, and the dominant pattern in contemporary multilingual
judge work), no such relation exists. The negative is sampled from a
distribution, not derived from the gold, so there is no transformation
to invert and no diff to take. The fault we document is exactly the one
this design cannot mechanize: a generated hallucination that is
\emph{too short} is, on the surface, indistinguishable from a generated
hallucination that \emph{is} short. Deciding which one occurred (whether
the model produced a degenerate but legitimate hallucination, or whether
it was prevented from finishing) requires reading the text, which is
precisely the expensive semantic act an oracle is meant to spare.
Aggregate statistics cannot help, because they were never looking at the
right object: they summarize judge behavior across items, not the
integrity of the items themselves.

\textbf{A positive control.} The claim that the asymmetry is about
oracles, not about bug frequency, is testable, and we tested it. We drew
fifty Turkish sentences containing numerals from the same wiki40b stream
used in \hyperlink{sec:4.1}{Section~4.1}, and built a minimal
mechanical-perturbation pipeline that changes the first numeral in each
sentence. We then injected the closest mechanical analogue of our fault:
via per-item independent Bernoulli draws at a 30\% rate (frozen by seed
42), the substitution is silently skipped, a no-op that emits a negative
identical to its gold. (The realized count was 20 no-ops, 40\% of the 50
items, about 1.5 SD above the expected 15, ordinary binomial fluctuation
rather than a separate effect.) This is the same class of failure (a
silent degeneration producing a useless negative) in a design that
\emph{does} carry an oracle. The oracle, a gold-to-negative string
comparison, flagged every no-op and confirmed every genuine
perturbation, at 50/50 accuracy, with zero API calls, zero human
reading, and O(N) cost. The same fault class that required four rounds
of robustness checking plus a manual read to surface in our oracle-less
design was caught mechanically and instantly here. We do not claim
mechanical-perturbation scripts are bug-free; we claim their bugs are
\emph{findable}, and that is the entire difference (Table 3).

\begin{table}[H]
\centering
\caption{The two corpus-construction designs, separated by whether item integrity is mechanically checkable.}
\begin{tabular}{@{}
  >{\raggedright\arraybackslash}p{(\columnwidth - 6\tabcolsep) * \real{0.2500}}
  >{\raggedright\arraybackslash}p{(\columnwidth - 6\tabcolsep) * \real{0.2500}}
  >{\raggedright\arraybackslash}p{(\columnwidth - 6\tabcolsep) * \real{0.2500}}
  >{\raggedright\arraybackslash}p{(\columnwidth - 6\tabcolsep) * \real{0.2500}}@{}}
\toprule
\begin{minipage}[b]{\linewidth}\raggedright
Design
\end{minipage} & \begin{minipage}[b]{\linewidth}\raggedright
Item-level oracle?
\end{minipage} & \begin{minipage}[b]{\linewidth}\raggedright
Cost to surface a silent degeneration fault
\end{minipage} & \begin{minipage}[b]{\linewidth}\raggedright
Representative work
\end{minipage} \\
\midrule
Mechanical perturbation & \textbf{Yes}, gold-to-negative relation
checked by string diff (metamorphic) & O(N), no semantic judgment &
BabelJudge (2026) \\
LLM-generated negatives & \textbf{No}, negative sampled not derived;
``real vs truncated'' is a semantic call & requires human reading
(\hyperlink{sec:5}{Section~5}) & this work; other LLM-generated-negative
corpora (CalibJudge, 2026, cited as a candidate, unverified) \\
\bottomrule
\end{tabular}
\end{table}

\textbf{What the framework does and does not claim.} It claims that
item-level \emph{auditability} is a property of the generation design,
and that the designs divide cleanly along the oracle line. It does
\emph{not} claim mechanical perturbation is otherwise superior: a
rule-built negative can be distributionally unnatural, can cover only
the perturbation types its authors implemented, and can encode its own
annotator biases into what counts as ``wrong.'' Those are real
limitations, on a different axis (naturalness and coverage of the
negative space) than the one we isolate (whether item integrity is
mechanically checkable). It also does not claim that our one
positive-control fault type exhausts the class; a stronger version of
this argument would run the same audit-injection procedure across
multiple fault types and multiple oracle-bearing corpora, and we mark
that as future work rather than assert it.

\textbf{Why this matters.} Naming the asymmetry as an oracle problem
does two things. First, it relocates the contribution: the paper is no
longer ``we found a bug in our pipeline and invented a taxonomy for
it,'' but ``we exhibit an existence proof that, in an oracle-less
generation design, the standard aggregate-robustness toolkit (larger
samples, replication, a converging mechanistic account, a controlled
manipulation) is \emph{by construction} unable to surface stimulus-level
faults, and we show the same fault class is mechanically catchable in an
oracle-bearing design.'' Second, it connects the finding to an
established literature: metamorphic testing and the oracle problem give
reviewers a vocabulary and a precedent for the claim, rather than asking
them to accept a bespoke taxonomy on faith. The protocol of
\hyperlink{sec:5}{Section~5} is then not an ad hoc checklist but the
predictable consequence of working in an oracle-less regime: when no
mechanical check exists, the only remaining check is the expensive one,
and the only question is whether the analyst ran it before computing
statistics.

\hypertarget{sec:5}{%
\section{5. A Validation Protocol for the Oracle-less
Regime}\label{sec:5}}

The four robustness procedures of \hyperlink{sec:4.1}{Section~4.1}
failed to surface the fault not because they were applied carelessly but
because none of them inspects the items themselves. The check that did
work, manual reading, is the one this section prescribes. We present it
as a cheap, bounded procedure an analyst can run before computing any
aggregate statistic, and we derive its thresholds from our own case
rather than asserting them as universal.

\textbf{The protocol.} Before any aggregate measurement, for every
distinct generation condition in the study (each producer, language, and
prompt template):

\begin{enumerate}
\def\labelenumi{\arabic{enumi}.}
\item
  \emph{Read.} Manually read fifteen to twenty raw generated items in
  full. The two endpoints are the two ends of one calculation, not two
  independent arguments. The sample size needed to see at least one
  faulty item with confidence C at fault rate p is n ≥ ln(1−C)/ln(1−p),
  which for C=99\% gives n=7 at a 50\% rate, n=13 at 30\%, n=21 at 20\%,
  and n=44 at 10\%. We use 30\% as the reference rate, borrowing the
  30\% target rate from the positive control of
  \hyperlink{sec:4.3}{Section~4.3} as an illustrative lower-bound
  scenario; our own fault ran at 100\%, so 30\% is conservative. At that
  rate fifteen items already clear 99\% confidence, since thirteen
  suffice, and twenty pushes the margin to about 99.9\%, or holds it
  near 99\% if the true fault rate is lower, around 20\%. Fifteen is
  therefore a floor that meets the bar at a 30\% rate, and twenty is a
  margin against lower or more diffuse fault rates.
\item
  \emph{Measure length.} Report the mean word count of the generated
  field per condition. In our case the corrupted and corrected pipelines
  produced distributions that barely overlapped (roughly two words
  versus twenty to forty-seven); a length that sits far from what the
  generation prompt was asked to produce is the first mechanical flag.
\item
  \emph{Measure degeneration.} Report two rates: the fraction of items
  whose generated field is identical to its source factual field, and
  the fraction that begin with a conversational opener or consist of a
  fragment shorter than three words. In our fault these were 12.2\% and
  81.0\% respectively; either, reported alongside a result, tells a
  reader whether the items the statistics summarize were well-formed at
  all.
\end{enumerate}

\textbf{Where the thresholds come from, and where they do not.} The
specific numbers above, fifteen items and a particular length gap, are
not universal. They are the values our case happened to exhibit, and we
offer them as a worked example, not a calibrated standard. What we do
claim generalizes is the shape of the diagnostic, not its thresholds: a
generation fault of the class we document leaves a bimodal signature,
with the affected condition's length and degeneration rate separating
sharply from the unaffected ones, and an analyst working in a different
task or language should look for that separation, not for our exact
numbers. A task whose well-formed answers are themselves short, or whose
faults manifest as subtle paraphrase rather than truncation, will have
different length and degeneration baselines, and the protocol's job
there is to establish those baselines before measuring bias, not to
import ours.

\textbf{What the protocol catches, and what it leaves.} The protocol is
designed to catch item-level degeneration, the fault class of
\hyperlink{sec:4.1}{Section~4.1}, before it contaminates a downstream
measurement. It is not designed to detect judge bias itself: a corpus of
perfectly well-formed items can still measure a biased judge, and
confirming item integrity does not validate the bias estimate. It is
also not a substitute for an oracle. An oracle-bearing design catches
the same fault class mechanically, at O(N), with no human reading, as
\hyperlink{sec:4.3}{Section~4.3} demonstrates; the protocol here is the
stopgap for designs that cannot carry an oracle, which under our
argument is most contemporary multilingual LLM-as-judge work. Its virtue
is not that it is powerful but that it is cheap, that it looks at the
right object, and that the alternative to running it is not a better
statistical check but no check at all. The measurement comes with one
decision rule: a condition whose degeneration rate rises above single
digits should be resolved (the pipeline fixed, or the condition
excluded) before it enters the statistics, not reported with a suspicion
noted afterward. The single-digit cut is, like the thresholds above, a
value our case implies rather than a universal line; the principle
behind it is that a condition whose items are not well-formed cannot
support a measurement, and that should be settled before aggregation,
not after.

\hypertarget{discussion}{%
\section{6. Discussion}\label{discussion}}

\textbf{What the case establishes, and what it does not.} We have used a
single pipeline fault to make a structural argument, and the honest way
to state that argument is narrow. We have shown that, in one concrete
and A/B-verified instance, a stimulus-level fault in an oracle-less
generation design survived every standard aggregate-robustness procedure
we applied to it, larger samples and replication and a converging
mechanistic account and a controlled manipulation included, and was
exposed only by reading the raw items by hand
(\hyperlink{sec:4.1}{Section~4.1}). We have shown the same fault class,
injected into an oracle-bearing design, was caught mechanically and at
no cost (\hyperlink{sec:4.3}{Section~4.3}). From these two
demonstrations we draw one conclusion we think is secure: whether a
generation design admits a cheap item-level check is a property of the
design, and that property governs the catchability of this fault class
more than the frequency of the bugs themselves does. We do not claim to
have shown that every oracle-less corpus is silently broken, nor that
every oracle-bearing corpus is clean. We claim that the two designs put
the analyst in structurally different positions with respect to finding
out.

\textbf{Why the oracle framing earns its keep.} We could have stopped at
``read your raw generations,'' which is the practical advice and the one
we expect most readers to act on. We chose to frame the finding through
the test oracle problem (\hyperlink{ref:barr}{Barr et al., 2015}) and
metamorphic testing (\hyperlink{ref:chen1998}{Chen et al., 1998}) for
two reasons that are not cosmetic. First, it relocates the result from a
personal cautionary tale into a class: the four failed robustness checks
did not fail because we were careless, they failed because none of them
is an oracle, and the absence of an oracle is a named, studied condition
in software engineering with its own body of mitigations. Second, it
makes the positive control legible: the reason our perturbation pipeline
caught the injected fault is not that we wrote a clever audit, but that
the perturbation itself is a metamorphic relation between gold and
negative, and checking a relation is cheaper than judging a free-form
generation. A reader who disagrees with our specific protocol can still
agree with the framing, and then ask the productive question, which is
not ``did they read enough items'' but ``does my design carry an oracle,
and if not, have I paid the cost an oracle-less design demands.''

\textbf{The asymmetry is about findability, not quality, and it implies
a choice rather than a verdict.} We do not promote mechanical
perturbation as the better design overall. When a research question
genuinely requires fluent, natural negatives, a rule-built answer is the
wrong tool, and nothing in our argument says to move to it. What the
findability asymmetry does is price the alternative correctly. The
cheap, mechanical item-level check does not exist in the oracle-less
regime, so the analyst who stays in it owes the reader the one check
that does: the manual read of \hyperlink{sec:5}{Section~5}, performed
before any aggregate number is computed. Mechanical perturbation is the
alternative for the analyst whose question tolerates less natural
negatives and who would rather carry the item-level oracle as insurance
than pay for the manual read. The choice between the two is
research-question-dependent and entirely legitimate either way. What is
not optional, in either design, is knowing which regime you are in and
paying its cost up front.

\textbf{Limitations.} Several scoping conditions bound the argument, and
we state them as limits rather than as weaknesses to be talked around.
The existence proof is a single pipeline and a single fault (a shared
decoding budget); the positive control exercises one fault type (a
silent no-op) in one oracle-bearing corpus. We have not shown that the
catchability asymmetry holds across fault types, across domains, or
across oracle-bearing corpora, and the natural strengthening of our
claim, a sweep over fault classes and over both design families, is
explicitly future work. The judge set is three models from three
families, the producer comparison is one pair on one language, and the
source domain is a single encyclopedic stream. Each of these could be
widened, and widening them would either reinforce the argument (if the
asymmetry reproduces) or refine it (if it does not); we would welcome
either result over the present single-instance generality. Finally, the
entire measurement operates on a binary faithfulness classification,
factual-versus-hallucinated; how a generation fault of this class
propagates into preference or quality-comparison judgments, where there
is no correct reference answer to compare against, is a genuinely
different question and one this paper does not address.

\textbf{Consistency at our own expense: the English reference.} A reader
who applies our protocol to our own evidence will notice an asymmetry we
are obliged to name rather than hide. We prescribed a
fifteen-to-twenty-item manual read per generation condition and derived
our length threshold from our Turkish data. HaluEval, our English
reference, is a reputable and established benchmark, and we treated its
reputation as grounds for not re-auditing its individual items against
the standard we now recommend. We did not, in other words, run our own
protocol on our own English reference. We state this plainly for three
reasons. It is honest. It is exactly the kind of default-trust that the
oracle-less regime encourages and that our argument warns against, which
means our corpus is itself an instance of the problem and not only its
demonstration. And it leaves a concrete open task: an item-level audit
of HaluEval under the \hyperlink{sec:5}{Section~5} protocol would either
confirm its short-regime style scores as genuine or surface a
stimulus-level issue we have not seen, and both outcomes would be
informative. We do not assert HaluEval is clean; we assert we have not
checked, and we name that as the predictable consequence of working in a
regime where checking is left to the analyst.

\textbf{The general lesson.} Synthetic-data generation pipelines are
infrastructure, and infrastructure that is unaudited will, sometimes,
fail silently. The lesson we draw is narrower than ``audit everything''
and stronger than ``we made a mistake.'' It is that, in the oracle-less
design that now dominates multilingual LLM-as-judge work, the integrity
of the stimuli is not a property the statistics can recover, and the
only mechanism that bounds it is the analyst's own decision to look at
the raw items before computing anything. The community-level implication
is not a new benchmark or a new metric. It is the cheaper and less
glamorous practice of reporting, alongside every aggregate bias number,
the smallest possible fact about the inputs that produced it: how long
the generated items were, how many were degenerate, and whether anyone
read them. A field that made that practice routine would have caught our
fault on the first round, and so would we.

\hypertarget{conclusion}{%
\section{7. Conclusion}\label{conclusion}}

We reported a single, well-localized fault in a synthetic LLM-as-judge
corpus, a shared decoding budget that silently truncated one producer's
generations, and we used it to make a structural argument rather than a
cautionary one. The argument is that, in the generation design now
dominant in multilingual judge work, the standard aggregate-robustness
toolkit is not equipped to detect a fault in the stimuli themselves,
because it operates downstream of stimulus integrity and was never
looking at it. In our case four independent checks, larger samples and
replication and a converging mechanistic account and a controlled
manipulation, each made the fabricated effect more convincing, and only
manual reading of the raw items surfaced it. We do not claim our four
checks were performed carelessly; we claim they are the wrong instrument
for this fault class, by construction.

The reason is not personal but structural, and it has a name. A corpus
whose negative examples are LLM-generated carries no mechanical
item-level oracle: a generated hallucination that is degenerate is
surface-identical to one that is merely short, and telling them apart is
the expensive semantic act an oracle exists to spare. A corpus whose
negatives are derived from a gold answer by deterministic perturbation
carries such an oracle for free, as a gold-to-negative relation
checkable by a string comparison. This is the test oracle problem of
software engineering (\hyperlink{ref:barr}{Barr et al., 2015}) and the
logic of metamorphic testing
(\hyperlink{ref:chen1998}{Chen et al., 1998}), applied to corpus
construction rather than to the judge. Our positive control made the
asymmetry concrete: the same fault class that escaped four rounds of
robustness checking in our oracle-less design was caught mechanically
and instantly, at no API or human cost, in an oracle-bearing one. The
asymmetry is about findability, not about bug frequency or design
quality.

What we offer the analyst who must work in the oracle-less regime is
accordingly modest and specific: a short manual reading of fifteen to
twenty raw items per generation condition, a reported length and
degeneration rate, and a decision rule that excludes a condition whose
items are not well-formed before it enters the statistics. The
thresholds come from our case, not from a universal standard; what
generalizes is the diagnostic shape and the obligation to look at the
inputs before measuring what they produce. We are equally deliberate
about what the paper does not claim. It does not discover a new judge
bias: the one our corrupted pipeline dramatized was an artifact, the one
it spared is a known formatting preference. It does not census affected
corpora: we have audited our own pipeline alone and name candidate
members of the class without verifying them. And it does not rank
generation designs: mechanical perturbation is not better, its faults
are merely findable. Within that scope the conclusion is narrow and, we
think, durable. In synthetic judge-bias research the integrity of the
stimuli bounds the integrity of every measurement built on them; whether
that bound is cheaply checkable is a property of the generation design;
and the default design leaves the check to the analyst's discipline. The
contribution is small, and we mean it to be: one verified case, one
positive control, and one cheap protocol, each of which an analyst can
check against their own pipeline before trusting a single aggregate
number it contains.

\hypertarget{references}{%
\section{References}\label{references}}

\hypertarget{ref:alansari}{}

Alansari, A., \& Luqman, H. (2026). Large language models hallucination:
A comprehensive survey. \emph{Computer Science Review}, 61, Article
100970.
\href{https://doi.org/10.1016/j.cosrev.2026.100970}{doi:10.1016/j.cosrev.2026.100970}

\hypertarget{ref:barr}{}

Barr, E. T., Harman, M., McMinn, P., Shahbaz, M., \& Yoo, S. (2015). The
oracle problem in software testing: A survey. \emph{IEEE Transactions on
Software Engineering}, 41(5), 507--525.
\href{https://doi.org/10.1109/TSE.2014.2372785}{doi:10.1109/TSE.2014.2372785}

\hypertarget{ref:chen1998}{}

Chen, T. Y., Cheung, S. C., \& Yiu, S. M. (1998). \emph{Metamorphic
testing: A new approach for generating next test cases} (Technical
Report HKUST-CS98-01). Hong Kong University of Science and Technology.
\href{https://arxiv.org/abs/2002.12543}{arXiv:2002.12543}

\hypertarget{ref:chen2018}{}

Chen, T. Y., Kuo, F.-C., Liu, H., Poon, P.-L., Towey, D., Tse, T. H., \&
Zhou, Z. Q. (2018). Metamorphic testing: A review of challenges and
opportunities. \emph{ACM Computing Surveys}, 51(1), Article 4.
\href{https://doi.org/10.1145/3143561}{doi:10.1145/3143561}

\hypertarget{ref:glm4}{}

Team GLM. (2024). ChatGLM: A family of large language models from
GLM-130B to GLM-4 all tools. arXiv preprint.
\href{https://arxiv.org/abs/2406.12793}{arXiv:2406.12793}

\hypertarget{ref:glm5}{}

GLM-5 Team. (2026). GLM-5: From vibe coding to agentic engineering.
arXiv preprint.
\href{https://arxiv.org/abs/2602.15763}{arXiv:2602.15763}

\hypertarget{ref:gu2025}{}

Gu, J., Jiang, X., Shi, Z., Tan, H., Zhai, X., Xu, C., Li, W., Shen, Y.,
Ma, S., Liu, H., Wang, S., Zhang, K., Wang, Y., Gao, W., Ni, L., \& Guo,
J. (2025). A survey on LLM-as-a-judge. \emph{The Innovation}, 7(6),
Article 101253.
\href{https://doi.org/10.1016/j.xinn.2025.101253}{doi:10.1016/j.xinn.2025.101253}

\hypertarget{ref:li2026}{}

Li, D., Sun, B., Huang, W., Zhong, T., Jiang, Z., Han, J., Zhang, X.,
Wang, Z., \& Liu, T. (2026). Preference leakage: A contamination problem
in LLM-as-a-judge. In \emph{International Conference on Learning
Representations (ICLR)}.
\href{https://arxiv.org/abs/2502.01534}{arXiv:2502.01534}

\hypertarget{ref:li2023}{}

Li, J., Cheng, X., Zhao, W. X., Nie, J.-Y., \& Wen, J.-R. (2023).
HaluEval: A large-scale hallucination evaluation benchmark for large
language models. In \emph{Proceedings of EMNLP 2023}.
\href{https://doi.org/10.18653/v1/2023.emnlp-main.397}{doi:10.18653/v1/2023.emnlp-main.397}
\textbar{} \href{https://arxiv.org/abs/2305.11747}{arXiv:2305.11747}

\hypertarget{ref:ma}{}

Ma, C., Zhang, E., Zhao, Y., Liu, W., Jia, Y., Qing, P., Shi, L., Cohan,
A., Yan, Y., \& Vosoughi, S. (2025). Judging with many minds: Do more
perspectives mean less prejudice? On bias amplification and resistance
in multi-agent based LLM-as-judge. In \emph{Findings of EMNLP 2025}
(pp.~17356--17392).
\href{https://doi.org/10.18653/v1/2025.findings-emnlp.941}{doi:10.18653/v1/2025.findings-emnlp.941}

\hypertarget{ref:petrov}{}

Petrov, A., La Malfa, E., Torr, P., \& Bibi, A. (2023). Language model
tokenizers introduce unfairness between languages. In \emph{Advances in
Neural Information Processing Systems (NeurIPS 2023)}.
\href{https://arxiv.org/abs/2305.15425}{arXiv:2305.15425}

\hypertarget{ref:shi}{}

Shi, L., Ma, C., Liang, W., Diao, X., Ma, W., \& Vosoughi, S. (2025).
Judging the judges: A systematic study of position bias in
LLM-as-a-judge. In \emph{Proceedings of IJCNLP-AACL 2025}
(pp.~292--314).
\href{https://doi.org/10.18653/v1/2025.ijcnlp-long.18}{doi:10.18653/v1/2025.ijcnlp-long.18}

\hypertarget{ref:shreyas}{}

Shreyas KC. (2026). BabelJudge: Measuring LLM-as-a-judge reliability
across languages and agent trajectories. arXiv preprint.
\href{https://arxiv.org/abs/2606.22329}{arXiv:2606.22329}

\hypertarget{ref:soumik}{}

Soumik, S. K. (2026). Judging the judges: A systematic evaluation of
bias mitigation strategies in LLM-as-a-judge pipelines.
\emph{Transactions on Machine Learning Research} (under review;
OpenReview QF4lAmG4zc).
\href{https://arxiv.org/abs/2604.23178}{arXiv:2604.23178}

\hypertarget{ref:verga}{}

Verga, P., Hofstätter, S., Althammer, S., Su, Y., Piktus, A.,
Arkhangorodsky, A., Xu, M., White, N., \& Lewis, P. (2024). Replacing
judges with juries: Evaluating LLM generations with a panel of diverse
models. arXiv preprint.
\href{https://arxiv.org/abs/2404.18796}{arXiv:2404.18796}

\hypertarget{ref:wen}{}

Wen, C., Zhu, A., Long, R., Huang, H., Jiang, J., \& Lee, C. S. (2026).
CalibJudge: Calibrated LLM-as-a-judge for multilingual RAG with
uncertainty-aware scoring. Preprint. Cited as a candidate member of the
LLM-generated-negative design class; generation method not verified.
\href{https://doi.org/10.20944/preprints202603.1324.v1}{doi:10.20944/preprints202603.1324.v1}

\hypertarget{ref:zhang}{}

Zhang, Y., Wang, C., Wu, L., Yu, W., Wang, Y., Bao, G., \& Tang, J.
(2026). UDA: Unsupervised debiasing alignment for pair-wise
LLM-as-a-judge. \emph{Proceedings of the AAAI Conference on Artificial
Intelligence}, 40(41), 34854--34861.
\href{https://doi.org/10.1609/aaai.v40i41.40788}{doi:10.1609/aaai.v40i41.40788}

\hypertarget{ref:wiki40b}{}

wiki40b. (2020). \emph{Multilingual language model benchmark}. Dataset.
\url{https://huggingface.co/datasets/google/wiki40b}

\end{document}